# A Proposed Decision Support System/Expert System for Guiding Fresh Students in Selecting a Faculty in Gomal University, Pakistan

Muhammad Zaheer Aslam*, Nasimullah, Abdur Rashid Khan

Gomal University DlKhan.Pakistan

*zaheer.aslam@yahoo.com

**Abstract**

This paper presents the design and development of a proposed rule based Decision Support System that will help students in selecting the best suitable faculty/major decision while taking admission in Gomal University, Dera Ismail Khan, Pakistan. The basic idea of our approach is to design a model for testing and measuring the student capabilities like intelligence, understanding, comprehension, mathematical concepts plus his/her past academic record plus his/her intelligence level , and applying the module results to a rule-based decision support system to determine the compatibility of those capabilities with the available faculties/majors in Gomal University. The result is shown as a list of suggested faculties/majors with the student capabilities and abilities.

## 1. Introduction

When students complete their pre-university education, they take admission in university in a particular field/area of study for their bachelor studies. This is a very critical stage for them because their whole professional career depends on it. So a best selection of Major is very important for them at this stage. We know that different fields of studies have difference among them and they have different basic requirements. Similarly all students don't have the same level of intelligence, understanding, comprehension, mathematical concepts, his/her own interest in a field and his/her past academic record. So every field cant suit him/her, a special field will be good for him/her in achieving good results for future. There is also another factor affecting taking admission is, the basic criterion of every department which is also different within a university or from university to university. So we can say that only personal desire is not enough alone to make him/her successful in it and not his/her desire will work successfully for him/her. Some people lose their money and time in a field that is not compatible with their capabilities and abilities. The past statistical evidence have shown that a lot of students fail in their university studies even though, they receive all the family support and they are not weak in their studies but as they did not chose the right faculty/major that was compatible with their capabilities and abilities Many students choose a university faculty/major because it has a good social reputation or their friends have chosen it. The student does not know the extent of their real capabilities and abilities, and they do not have adequate capacity to learn capabilities needed for each faculty/major. Decision Support systems can do this task through the provision of some of the questions asked in various fields to measure student capabilities in these areas and the intelligence level. Decision support system helps in making effective decisions as it allows us to do only right things. We have proposed a rule-based decision support system working like an expert system that contains a general rule-base and an inference engine. The inference engine retrieves rules from the rule-base to solve new problems based on the rules for similar problems stored in the rule-based system. In this way, a rule-based DSS can exhibit humanlike performance in that knowledge that can seemingly be acquired through experience [4]. We have developed our rule base DSS using CLIPS(C Language Integrated Production System) [7] language. CLIPS is a multi-paradigm programming language that provides support for rule-based, object-oriented, and procedural programming. CLIPS is a forward chaining, rule based production-system language, which is based on the RETE algorithm for pattern-matching. A command-line interpreter is the default interface for CLIPS language [2]. A few studies and systems have been published which advises the students in choosing a university. MyMajors [3] is a website which gives the student online advisement report on the suitable universities. The paper is organized as follows: section 2 describes the general idea of our rule base DSS and problem identification, section 3 presents knowledge base representation and CLIPS rules, section 4 presents student capabilities and evaluation test and intelligence level test while evaluation test and the result of the DSS is in section 5, and section 6 concludes the paper and outline direction for future research. At the end section 7 gives the references.

**2. Problem identification**

Different people have different specific mental abilities, and every university faculty/major needs special abilities and capabilities in their departments. So we need to measure the student mental abilities accurately and compare then with the university faculty/major required abilities and suggest a suitable faculty/major for the student so that he may be academically successful. For this purpose we need a decision support system/Expert System which will get student general background information like: student name, academic record, academic type and intermediate education passing year. Secondly the student's abilities and capabilities should also be checked through online abilities test to, thirdly we test the student's intelligence level by taking intelligence test. English language is the default language used in this test module. Applying the results of the test to our knowledge based system is the fourth step. Finally our DSS determines which faculty/major in general can be suitable for the student.

**3. Knowledge Base Representaion**

Main sources for knowledge for developing this DSS/ES are human academic experts of Gomal university, the admission criteria of gomal University, the past record of the students and their future result. We have collected all the required criterions, abilities and capabilities for each faculty/major in Gomal University. We converted the knowledge into facts and rules in CLIPS syntax, and store them in the knowledge base of the CLIPS language. CLIPS is suitable for forward reasoning and can be used easily to build the rules and facts, so that is why we have selected the CLIPS language.

Furthermore, we captured all general important background information about the student in CLIPS fact templates which are just like the structures used in C language. The following sample code shows a student structure for CLIPS template.

```
(deftemplate student "Student Data"
(slot stdid)
(slot name)
(slot age)
(slot academic-per)
(slot academic -type)
(slot HSSC-year)
(slot int-test-per)
(slot ability-test-eng-per)
(slot ability-test-phy-per)
(slot ability-test-che-per)
(slot ability-test-cs-per)
(slot ability-test-math-per)
(slot ability-test-bio-per)
)
```

The rule is represented in the syntax of "If conditions then actions ". That means when the conditions are satisfied then the actions are carried out. Here is an example from our decision support system using CLIPS syntax rules for the requirements of the faculty of Engineering:

```
(defrule fo-Mathematics
(student (stdid ?stdid)
(academic-per ?academic-per)
(academic-type ?academic-type)
(HSSC-year ?HSSC-year)
(int-test-per ?int-test-per)
(ability-test-math-per ? ability-test-math-per)
(ability-test-eng-per ?ability-test-eng-per)
(ability-test-phy-per ?ability-test-phy-per)
(ability-test-che-per ?ability-test-che-per)
)

(test (>= ?academic-per 60))
(test (eq ?academic-type Science))
(test (>= ?HSSC-year 2009))
=>
 (assert (faculty-of-Mathematics ?stdid))
```

```
(printout fddatao "[Mathematics]" crlf
"Accepted=TRUE" crlf)

(if
(and (>= ?int-test-per 80) (>= ?ability-test-math-per 80) (>=
?ability-test-eng-per 60) (>= ?ability-test-phy-per 60)
(>= ?ability-test-che-per 60))
then
(printout fddatao "Recommended=TRUE" crlf
crlf)
else (printout fddatao "Recommended=FALSE"
crlf crlf))
)
```

**4. Tests**

**We have divided the tests into two portions.**

## 1)   Capabilities & Abilities Test

## 2)   Intelligence Test

**4.1 Capabilities & Abilities Test**

Capability test evaluates student capabilities and abilities in many fields like Science, Mathematics, History, Geography, English language and other capabilities which he has studied before during his/her academic. We used English language for test modules. CLIPS language does not support graphical interfaces; but can be integrated with other high level languages like visual basic easily. We developed the test modules using Microsoft visual basic and then the user result at the end of the test is converted into text file which is then opened by the CLIPS DSS for reading the data.

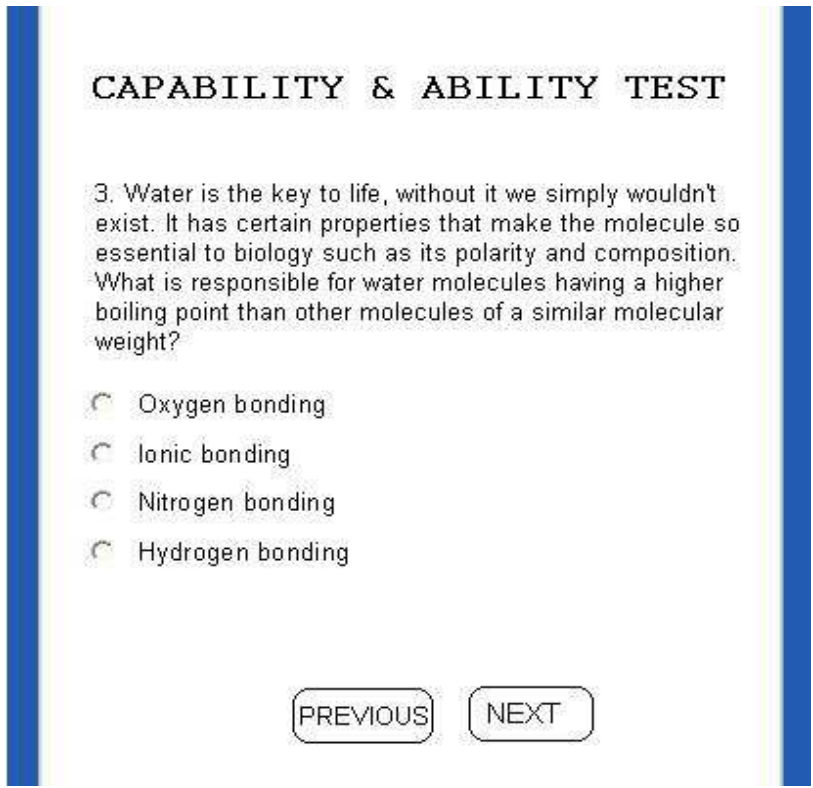

Figure (1): sample question screen from a capability & ability module test.

First we enter some important background information about the student like his name, academic score, academic type, year of intermediate, age, etc. Then a timed limited module test consisting of 100 questions begins which contains 20 questions each for English, Mathematics, Physics, Chemistry, Computer Science / Biology. Figure1 shows a sample question screen from a capabilities module test. As soon as the abilities test ends, our DSS have all information about the student capabilities and abilities captured. This information then is converted into facts in our CLIPS Decision support system. The following rule which is executed at start shows how the information is transformed into CLIPS facts:

```
(defrule readtextfiledata
(declare (salience 2000))
(initial-fact)
?factstd <- (initial-fact)
=>
(retract ?factstd)
(open "std-data-in.txt" fddatai "r")
(open "std-data-out.txt" fddatao "w")
```

```
(bind ?stdid (read fddatai))
(bind ?name (read fddatai))
(bind ?age (read fddatai))
(bind ?academic-per (read fddatai))
(bind ?academic-type (read fddatai))
(bind ?HSSC-year (read fddatai))
(bind ?int-test-per (read fddatai))
(bind ?ability-test-eng-per (read fddatai))
(bind ?ability-test-phy-per (read fddatai))
(bind ?ability-test-che-per (read fddatai))
(bind ?ability-test-cs-per (read fddatai))
(bind ?ability-test-math-per (read fddatai))
(bind ?ability-test-bio-per (read fddatai))

 (printout fdatao "[STUDENTINFO]" crlf "No="
?stdid crlf "name= " ?name crlf "Age=" ?age crlf "Academic Percentage=" ?academic-per
crlf "Academic Type=" ?academic-type crlf "HSSC-Year="
?HSSC-year crlf crlf)

(assert (student
(stdid ?stdid)
(name ?name)
(age ?age)
(academic-per ?academic-per)
(academic-type ?academic-type)
(HSSC-year ?HSSC-year)
(int-test-per ?int-test-per)
(ability-test-eng-per ?ability-test-eng-per)
(ability-test-phy-per ?ability-test-phy-per)
(ability-test-che-per ?ability-test-che-per)
(ability-test-cs-per ?ability-test-cs-per)
(ability-test-math-per ? ability-test-math-per)
(ability-test-bio-per ? ability-test-bio-per)
)
```

### 4.2 Intelligence Test

This test starts soon after the academics test ends. It evaluates student Intelligence. We have developed the intelligence test just like we developed the Capability test discussed in section 4. After ending the capability test, the intelligence test starts without break. It has a timed limited module test consisting of 50 questions. Figure2 shows a sample question screen from an intelligence module test. By the time the student finishes the intelligence test; our DSS have all information about the student intelligence captured. This information then is converted into facts in our CLIPS DSS system.

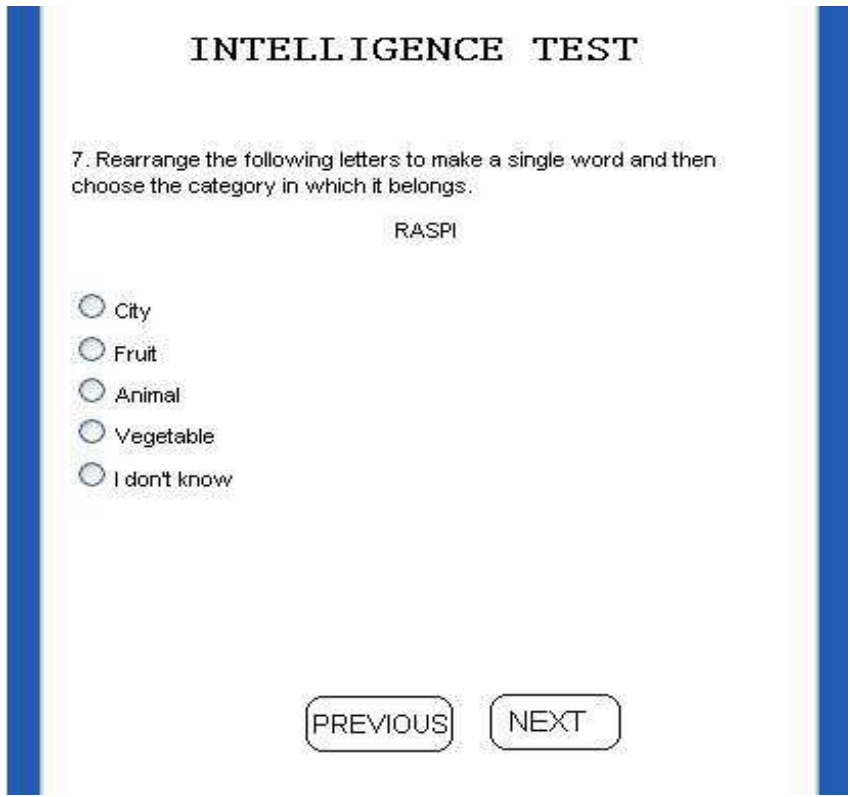

Figure (2): sample question screen from a intelligence module test.

## 5. Result Of The Expert System/Dss

At the end our decision support system makes recommendations for the student. The recommendation is made based on the student's background information, academic record, intelligence test results and the abilities module test results. Our DSS identify the most suitable faculty or major for the student based on his abilities and capabilities extracted from the test module results. These results may not be 100% accurate, because there may be several factors that may affect the student away from the scientific level, such as mental state and health situation during the evaluation test, which will affect negatively on the results.

## 6. Conclusion & Future Work

The paper we have presented gives the design and development of a proposed rule based decision support system to help the students to select the best suitable faculty or major based on their capabilities and abilities. Our DSS is a rule based system, and we used CLIPS language to store our knowledge base. Using abilities test and intelligence test and their past academic record, we can measure some student capabilities and abilities and determine which faculty/major is suitable for him/her. The fact is measuring abilities and capabilities of the student accurately are complex process. But, as more students being evaluated in the abilities test the more prospect realistic results is withdrawn. Our DSS definition rules can be made more customized and more criteria may be added to it for more data mined results. It can be extended to include other universities faculties and majors to be able to serve more students wishing to be enrolled in other universities and make the criterion customized for that university.

**7. References**

[1]. Wikipedia Encyclopedia ,
[2]. Zhengxin Chen, Computational Intelligence
for Decision Support, 1999, University of Nebraska, Omaha.
[3]. MyMajors, http://www.mymajors.com .
[4]. Y.-Q. Zhang and others , Comeutational Web Intelligence (Series in Machine Perception and Artificial Intelligence (Vol. 58)),2004 Georgia State University, Atlanta, Georgia, USA .
[5]. Azaab S., Abu Naser S., and Sulisel O.,2000. A proposed expert system for selecting exploratory factor analysis procedures, Journal of the college of education, 4(2):9-26.
[6]. Russell, S. and P. Norvig, 2002. Artificial Intelligence: A Modern Approach, Prentice Hall, Second Edition.
[7] "CLIPS language homepage developed by NASA" http://www.siliconvalleyone.com/clips.htm